\documentclass[twoside]{article}
\usepackage[accepted]{aistats2012}

\usepackage{times}
\usepackage{epsfig}
\usepackage{graphicx}
\usepackage{amsmath}
\usepackage{amssymb}
\usepackage{picins}
\usepackage[usenames]{color} 
\usepackage[USenglish]{babel}
\usepackage[ruled,vlined]{algorithm2e} 
\usepackage{picins} 
\usepackage{multirow}

\usepackage{bbm}

\usepackage{hyperref}

\definecolor{Blue}{rgb}{0,0,1}
\definecolor{Red}{rgb}{1,0,0}
\definecolor{Green}{rgb}{0,.8,0}
\definecolor{Magenta}{rgb}{1,0,1}

\definecolor{swap}{rgb}{0,0,1}
\definecolor{expand}{rgb}{0.44,0.44,0.44}
\definecolor{trws}{rgb}{0,1,1}
\definecolor{bp}{rgb}{1,0,0}
\definecolor{icm}{rgb}{0,1,0}
\definecolor{rlp}{rgb}{1,0,1}

\newcommand{\comment}[1]{}
\newcommand{\abs}[1]{\mbox{$\left|#1\right|$}}

\newcommand{\norm}[1]{\left\| #1 \right\|}
\newcommand{\uut}{\left[UU^T\right]_{ij}}

\newcommand{\TabCenter}[2]{\begin{minipage}[t]{#1}\centering #2\end{minipage}}
\newcommand{\etal}{\mbox{et~al.}}

\begin{document}


\author{Shai Bagon \quad \quad  Meirav Galun\\
Dept. of Computer Science and Applied Mathmatics\\
The Weizmann Institute of Scince\\
Rehovot 76100, Israel\\
http://www.wisdom.weizmann.ac.il/\{\texttildelow bagon, \texttildelow meirav\}
}

\twocolumn[
\aistatstitle{Optimizing Large Scale Correlation Clustering}
\aistatsauthor{ Shai Bagon \quad Meirav Galun }
\aistatsaddress{ Dept. of Computer Science and Applied Mathmatics\\
The Weizmann Institute of Scince\\
Rehovot 76100, Israel\\
http://www.wisdom.weizmann.ac.il/$\sim$\{bagon, meirav\} }
]

\begin{abstract}
Clustering is a fundamental task in unsupervised learning.
The focus of this paper is the Correlation Clustering functional which combines positive and negative affinities between the data points.
The contribution of this paper is two fold:
(i)~Provide a theoretic analysis of the functional.
(ii)~New optimization algorithms which can cope with large scale problems
($>100K$ variables) that are infeasible using existing methods.
Our theoretic analysis provides a probabilistic generative interpretation for the functional, and
justifies its intrinsic ``model-selection" capability.
Furthermore, we draw an analogy between optimizing this functional and the well known Potts energy minimization.
This analogy allows us to suggest several new optimization algorithms,
which exploit the intrinsic ``model-selection" capability of the functional
to automatically recover the underlying number of clusters.
We compare our algorithms to existing methods on both synthetic and real data.
In addition we suggest two new applications that are made possible by our algorithms:
unsupervised face identification and interactive multi-object segmentation by rough boundary delineation.
\end{abstract}
%

\section{\label{sec:intro}Introduction}
One of the fundamental tasks in unsupervised learning is clustering: grouping data points into coherent clusters.
In clustering of data points, two aspects of pair-wise affinities can be measured: (i)~{\em Attraction} (positive affinities), i.e.,  how likely are points $i$ and $j$  to be in the same cluster, and
(ii)~{\em Repulsion} (negative affinities), i.e., how likely are points $i$ and $j$ to be in different clusters.

Indeed, new approaches for clustering, recently presented by Yu and Shi \shortcite{Yu2001} and Bansal~\etal\ \shortcite{Bansal2004}, suggest to combine attraction and repulsion information.
Normalized cuts was extended by Yu and Shi \shortcite{Yu2001} to allow for negative affinities.
However, the resulting functional provides sub-optimal clustering results in the sense that it may lead to fragmentation of large homogeneous clusters.

The Correlation Clustering functional ({\bf CC}), proposed by Bansal~\etal\ \shortcite{Bansal2004}, tries to maximize the intra-cluster agreement (attraction)
and the inter-cluster disagreement (repulsion).
Contrary to many clustering objectives, the CC functional has an inherent
``model-selection" property allowing to {\em automatically} recover the underlying number of clusters \cite{Demaine2003}.

Optimizing CC is tightly related to many graph partitioning formulations \cite{Nowozin2009}, however it is
known to be NP-hard \cite{Bansal2004}.
Existing methods derive convex continuous relaxations to approximately optimize the CC functional.
However, these algorithms do not scale beyond a few hundreds of variables.
See for example, the works of \cite{Nowozin2009,Bagon2010,Vitaladevuni2010,Glasner2011}.

This work suggests a new perspective on the CC functional, showing its analogy to the known {\em Potts model}.
This new perspective allows us to leverage on recent advances in discrete optimization to propose new CC optimization algorithms.
We show that our algorithms scale to large number of variables ($>100K$), and in fact can tackle tasks that were {\bf infeasible in the past}, e.g., applying CC to pixel-level image segmentation.
In addition, we provide a {\em rigorous statistical interpretation} for the CC functional and justify its intrinsic model selection capability.
Our algorithms exploit this ``model selection" property to automatically recover the underlying number of clusters $k$.

The contributions of this paper are as follows:\\*
\noindent\textbullet~A rigorous probabilistic interpretation of the CC functional, justifying its intrinsic model selection capability.\\*
\noindent\textbullet~A new perspective to the functional, drawing analogy to the discrete Potts model.\\*
\noindent\textbullet~New large scale optimization algorithms, that stem from our new perspective.\\*
\noindent\textbullet~Our algorithms automatically recover the underlying number of clusters $k$.\\*
\noindent\textbullet~New applications in vision and graphics.

The first part of the paper (Sec.~\ref{sec:theory}) focuses on the theoretical probabilistic interpretation of the CC functional. The subsequent sections are dedicated to the second part of this work which concerns the optimization of the CC functional.
%

\section*{Correlation Clustering (CC) Functional}

Let $W\in\mathbb{R}^{n\times n}$ be an affinity matrix
combining attraction and repulsion: for $W_{ij}>0$ we say that $i$ and $j$ attract each other with certainty $\abs{W_{ij}}$, and for $W_{ij}<0$ we say that $i$ and $j$ repel each other with certainty $\abs{W_{ij}}$. Thus the sign of $W_{ij}$ tells us if the points attract or repel each other and the magnitude of $W_{ij}$ indicates our certainty.

Any $k$-way partition of $n$ points can be written as $U\in\left\{0,1\right\}^{n\times k}$ s.t. $U_{ic}=1$ iff point $i$ belongs to cluster $c$. $\sum_c U_{ic}=1\;\forall i$ ensure that every $i$ belongs to {\em exactly} one cluster.

The CC functional
maximizes the intra-cluster agreement \cite{Bansal2004}.
Given a matrix $W$\footnote{Note that $W$ may be sparse.
The ``missing" entries are simply assigned ``zero certainty" and therefore they do not affect the optimization.},
an optimal partition $U$ minimizes:
\begin{eqnarray}
\mathcal{E}_{CC}\left(U\right) &=&  - \sum_{ij} W_{ij}\sum_c U_{ic}U_{jc} \label{eq:CorrClust}\\
 & s.t. & U_{ic}\in \left\{0,1\right\} ,\; \sum_c U_{ic}=1 \nonumber
\end{eqnarray}
Note that $\sum_c U_{ic}U_{jc}$ equals 1 iff $i$ and $j$ belong to the same cluster.
For brevity, we will denote $\sum_c U_{ic}U_{jc}$ by $\uut$ from here on.

%

%
\section{\label{sec:theory}Probabilistic Interpretation}

This section provides a  probabilistic interpretation for the CC functional.
This interpretation allows us to provide a theoretic justification for the ``model selection" property of the CC functional.
Moreover, our analysis exposes the underlying implicit prior that this functional assumes.

We consider the following probabilistic generative model for matrix $W$.
Let $U$ be the true unobserved partition of $n$ points into clusters.
Assume that for some pairs of points $i,j$ we observe their pairwise similarity values $s_{ij}$.
These values are random realizations from either a distribution $f^+$ or $f^-$,
depending on whether points $i,j$ are in the same cluster or not. Namely,
\begin{eqnarray*}
p\left(s_{ij}=s\left|\uut=1\right.\right) &=& f^+\left(s\right) \\
p\left(s_{ij}=s\left|\uut=0\right.\right) &=& f^-\left(s\right)
\end{eqnarray*}

Assuming independency of the pairs, the likelihood of observing similarities $\left\{s_{ij}\right\}$ given a partition $U$ is then
\[
\mathcal{L}\left(\left\{s_{ij}\right\}\left|U\right.\right) = \prod_{ij} f^+\left(s_{ij}\right)^{\uut}\cdot f^-\left(s_{ij}\right)^{\left(1-\uut\right)}
\]
To infer a partition $U$ using this generative model we look at the posterior distribution:
\[
Pr\left(U\left|\left\{s_{ij}\right\}\right.\right)  \propto  \mathcal{L}\left(\left\{s_{ij}\right\}\left|U\right.\right) \cdot Pr\left(U\right)
\]
where $Pr\left(U\right)$ is a prior. Assuming {\em a uniform prior} over all partitions, i.e., $Pr\left(U\right)=const$, yields:
\[
Pr\left(U\left|\left\{s_{ij}\right\}\right.\right)  \propto \prod_{ij} f^+\left(s_{ij}\right)^{\uut}\cdot f^-\left(s_{ij}\right)^{\left(1-\uut\right)}
\]
Then, the negative logarithm of the posterior is given by
\begin{eqnarray*}
-\log Pr\left(U\left|\left\{s_{ij}\right\}\right.\right) &=& \hat{C} + \sum_{ij}\log f^+\left(s_{ij}\right)\uut \\
& & + \sum_{ij}\log f^-\left(s_{ij}\right)\left(1-\uut\right)
\end{eqnarray*}
where $\hat{C}$ is a constant not depending on $U$.

Interpreting the affinities as log odds ratios $W_{ij} = \log\left(\frac{f^+\left(s_{ij}\right)}{f^-\left(s_{ij}\right)}\right)$,
the posterior becomes
\begin{eqnarray}
-\log Pr\left(U\left|\left\{s_{ij}\right\}\right.\right) & = &
C - \sum_{ij} W_{ij}\uut
\label{eq:CorrClustPij}
\end{eqnarray}
That is, Eq.~(\ref{eq:CorrClustPij}) estimates the log-posterior of a partition $U$.
Therefore, a partition $U$ that minimizes Eq.~(\ref{eq:CorrClustPij}) is the {\bf MAP} (maximum a-posteriori) partition.
Since Eq.~(\ref{eq:CorrClust}) and Eq.~(\ref{eq:CorrClustPij}) differ only by a constant they share the same minimizer: the MAP partition.

\subsection{Recovering $k$ (a.k.a. ``model selection")}

We showed that the generative model underlying the CC functional has a {\em single} model for all partitions, regardless of $k$.
Therefore, optimizing the CC functional one need not select between different generative models to decide on the optimal $k$.
Comparing partitions with different $k$ is therefore straight forward and does not require an additional ``model complexity" term (such as BIC, 
MDL, 
etc.)

\begin{figure}
\centering
\parpic[r][r]{\includegraphics[width=.42\linewidth]{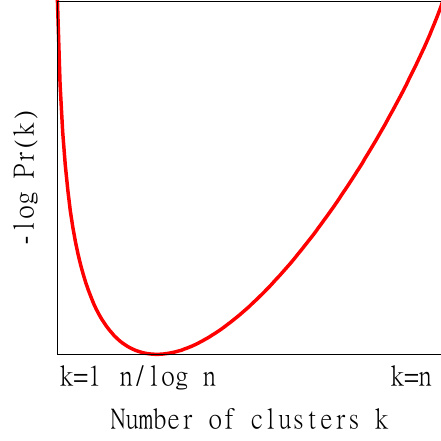}}
\caption{ {\bf Prior on the number of clusters $k$:}
{\em Graph shows $-\log Pr\left(k\right)$, for uniformly distributed $U$.
The induced prior on $k$ takes a non-trivial shape:
it assigns very low probability to the trivial solutions of $k=1$ and $k=n$,
while at the same time gives preference to partitions with non-trivial $k$.
The mode of this prior is when $U$ has roughly $\frac{n}{\log n}$ clusters.
}}
\label{fig:stirling}
\end{figure}

As described in the previous section the CC functional assumes a uniform prior over all partitions.
This uniform prior on $U$ induces a prior on the number of clusters $k$,
i.e., what is the a-priori probability of $U$ having $k$ clusters: $Pr\left(k\right)=Pr\left(U\mbox{ has $k$ clusters}\right)$.
We use Stirling numbers of the second kind \cite{Rennie1969} to compute this induced prior on $k$.
Fig~\ref{fig:stirling} shows the non-trivial shape of this induced prior on the number of clusters $k$.

%

%
\section{CC Optimization: Continuous Perspective}\label{sec:existing-cc-optimization}
After discussing the theory behind the CC functional and providing probabilistic justification to its model selection capability, we move to discuss methods and approaches to optimize this functional.
We begin with a brief glance at current state-of-the-art CC optimization algorithms.

Optimizing the correlation clustering functional (Eq.~(\ref{eq:CorrClust})) is NP-hard \cite{Bansal2004}.
Instead of solving {\bf directly} for a partition $U$, existing methods optimize {\bf indirectly} for the binary adjacency matrix $X=UU^T$, i.e., $X_{ij}=1$ iff $i$ and $j$ belong to the same cluster.
By introducing the binary adjacency matrix the quadratic objective (w.r.t. $U$): $-\sum_{ij} W_{ij}\uut$ becomes linear (w.r.t. $X$): $-\sum_{ij} W_{ij}X_{ij}$.
The connected components of $X$, after proper rounding, are the resulting clusters, and the number of clusters $k$ naturally emerges.
Indirect optimization methods must ascertain that the feasible set consists only of ``decomposable" $X$: $X=UU^T$.
This may be achieved either by posing semi-definite constraints on $X$ \cite{Vitaladevuni2010}, or by introducing large number of linear inequalities \cite{Demaine2003,Vitaladevuni2010}.
These methods take a continuous and convex relaxation approach to approximate the resulting functional.
This approach allows for nice theoretical properties due to the convex optimization at the cost of a very restricted scalability.

Solving for $X$ requires $\sim n^2$ variables instead of only \mbox{$\sim n$} when solving directly for $U$.
Therefore, these methods scale poorly with the number of variables $n$, and in fact, they cannot handle more than a few hundreds of variables.
In summary, these methods suffers from two drawbacks: (i)~recovering $U$ from $X$ is highly susceptible to noise and more importantly (ii)~it is {\em infeasible} to solve large scale problems by these methods.

%
\section{\label{sec:perspective}Our New Perspective on CC}

Existing methods view the CC optimization in the context of convex relaxation and build upon methods and approaches that are common practice in this field of continuous optimization.
We propose an alternative perspective to the CC optimization:
{\em viewing it as a discrete energy minimization}.
This new perspective allows us to build upon recent advances in discrete optimization and propose efficient and direct CC optimization algorithms.
More importantly, the resulting algorithms solve {\em directly} for $U$, and thus scales significantly better with the number of variables.

We now show how to cast the CC functional of Eq.(\ref{eq:CorrClust}) as a discrete pair-wise conditional random field (CRF) energy.
For ease of notation, we describe a partition $U$ using a labeling vector $L\in\left\{1,2\ldots\right\}^n$: $l_i = c$ iff  $U_{ic}=1$.
A general form of pair-wise CRF energy is $E\left(L\right)=\sum_i E_i\left(l_i\right) + \sum_{ij} E_{ij}\left(l_i, l_j\right)$ \cite{Veksler2002}. Discarding the unary term ($\sum_i E_i\left(l_i\right)$), and taking the pair-wise term to be $W_{ij}$ if $l_i \ne l_j$ we can re-write the CC functional as a CRF energy:
\begin{eqnarray}
\mathcal{E}_{CC}\left(L\right) = \sum_{ij} W_{ij} \cdot \mathbbm{1}_{\left[l_i\ne l_j\right]} \label{eq:CorrClustCRF}
\end{eqnarray}
This is a Potts model.
Optimizing the CC functional can now be interpreted as searching for a MAP assignment for the energy (\ref{eq:CorrClustCRF}).

The resulting Potts energy has three unique characteristics, each posing a challenge to the optimization process:\\*
\noindent(i)~{\bf Non sub-modular:} The energy is non sub-modular.
The notion of sub-modularity is the discrete analogue of convexity from continuous optimization \cite{Lovasz1983}.
Optimizing a non sub-modular energy is NP-hard, even for the binary case \cite{Rother2007}.\\*
\noindent(ii)~{\bf Unknown number of labels:}
Most CRF energies are defined for a fixed and known number of labels.
Thus, the search space is restricted to $L\in\left\{1,\ldots,k\right\}^n$.
When the number of labels $k$ is unknown the search space is by far larger and more complicated.\\*
\noindent(iii)~{\bf No unary term:} There is no unary term in the energy.
The unary term plays an important role in guiding the optimization process \cite{Szeliski2008}.
Moreover, a strong unary term is crucial when the energy in non sub-modular \cite{Rother2007}.

There exist examples of CRFs in the literature that share some of these characteristics (e.g., non sub-modular \cite{Rother2007,kolmogorov2005}, unknown number of labels \cite{Isack2011,Bleyer2010}).
Yet, to the best of our knowledge, no existing CRF exhibits all these three challenges at once.
More specifically, we are the first to handle non sub-modular energy that has no unary term.
Therefore, we cannot just use ``off-the-shelf" Potts optimization algorithms,
but rather modify and improve them to cope with the three challenges posed by the CC energy.

%

%
\begin{algorithm}[t!]
\caption{Expand-and-Explore\label{alg:a-expand}}
\DontPrintSemicolon
\SetKw{KwInit}{Init}
\SetKwFunction{KwExpand}{Expand}
\SetKwFunction{Weinberg}{$\mathcal{E}_{CC}$}

\KwIn{Affinity matrix $W\in\mathbb{R}^{n\times n}$}
\KwOut{Labeling vector $L\in\left\{1,2,\ldots\right\}^n$}
\BlankLine

\KwInit{$L_i\leftarrow 1$, $i=1,\ldots,n$}\tcp*[f]{initial labeling}\;
\Repeat{$L$ is unchanged}{
    \For{$\alpha\leftarrow1$ ; $\alpha\le\#L+1$ ; $\alpha++$}{
        $L \leftarrow $ \KwExpand{$\alpha$}\;
    } 
} 
\BlankLine

$\#L$ denotes the number of different labels in $L$.\;
\KwExpand{$\alpha$}: expanding $\alpha$ using QPBOI.\;
By letting $\alpha = \#L+1$ the algorithm ``expand" and explore an empty label. This may affect the number of labels $\#L$.\;
\end{algorithm}

\section{\label{sec:alg}Our Large Scale CC Optimization}

In this section we adapt known discrete energy minimization algorithms to cope with the three challenges posed by the CC energy.
We derive three CC optimization algorithms that stem from either large move making algorithms ($\alpha$-expand and $\alpha\beta$-swap \cite{Veksler2002}), or Iterated Conditional Modes (ICM) \cite{Besag1986}.
Our resulting algorithms scale gracefully with the number of variables $n$, and solve CC optimization problems that were infeasible in the past.
Furthermore, our algorithms take advantage of the intrinsic model selection capability of the CC functional (Sec.~\ref{sec:theory}) to robustly recover the underlying number of clusters.

\subsection{Improved large move making algorithms}
Boykov~\etal \shortcite{Veksler2002} introduced a very effective method for multi-label energy minimization that makes large search steps by iteratively solving binary sub-problems.
There are two large move making algorithms: $\alpha$-expand and $\alpha\beta$-swap that differ by the binary sub-problem they solve.
$\alpha$-expand consider for each variable whether it is better to retain its current label or flip it to label $\alpha$.
The binary step of $\alpha\beta$-swap involves only variables that are currently assigned to labels $\alpha$ or $\beta$, and consider whether it is better to retain their current label or switch to either $\alpha$ or $\beta$.
Defined for sub-modular energies, the binary step in these algorithms is solved using graph-cut.

We propose new optimization algorithms: {\em Expand-and-Explore} and {\em Swap-and-Explore}, inspired by $\alpha$-expand and $\alpha\beta$-swap, that can cope with the challenges of the CC energy.
(i)~For the binary step we use a solver that handles non sub-modular energies.
(ii)~We incorporate ``model selection" into the iterative search to recover the underlying number of clusters $k$.
(iii)~In the absence of unary term, a good initial labeling is provided to the non sub-modular binary solver.

Binary non sub-modular energies can be approximated by an extension of graph-cuts: QPBO \cite{Rother2007}.
When the binary energy is non sub-modular QPBO is not guaranteed to provide a labeling for all variables.
Instead, it outputs only a partial labeling. How many variables are labeled depends on the amount of non sub-modular pairs and the relative strength of the unary term for the specific energy.
When no unary term exists in the energy QPBO leaves most of the variables unlabeled. To circumvent this behavior we use the ``improve" extension of QPBO (denoted by QPBOI): This extension is capable of improving an initial labeling to find a labeling with lower energy \cite{Rother2007}.
In the context of expand and swap algorithms a natural initial labeling for the binary steps is to use the current labels of the variables and use QPBOI to improve on it, ensuring the energy does not increase during iterations.

To overcome the problem of finding the number of clusters $k$ our algorithms do not iterate over a fixed number of labels, but explore an ``empty" cluster in addition to the existing clusters in the current solution. Exploring an extra empty cluster allows the algorithms to optimize over all solutions with any number of clusters $k$. The fact that there is no unary term in the energy makes it straight forward to perform. Alg.~\ref{alg:a-expand} and Alg.~\ref{alg:ab-swap} presents our {\em Expand-and-Explore} and {\em Swap-and-Explore} algorithms in more detail.
\begin{algorithm}[t!]
\caption{Swap-and-Explore\label{alg:ab-swap}}
\DontPrintSemicolon
\SetKw{KwInit}{Init}
\SetKwFunction{KwSwap}{Swap}
\SetKwFunction{KwEnergy}{$\mathcal{E}_{CC}$}

\KwIn{Affinity matrix $W\in\mathbb{R}^{n\times n}$}
\KwOut{Labeling vector $L\in\left\{1,2,\ldots\right\}^n$}
\BlankLine

\KwInit{$L_i\leftarrow 1$, $i=1,\ldots,n$}\tcp*[f]{initial labeling}\;
\Repeat{$L$ is unchanged}{
    \For{$\alpha\leftarrow1$ ; $\alpha\le\#L$ ; $\alpha++$}{
        \For{$\beta\leftarrow\alpha$ ; $\beta\le\#L+1$ ; $\beta++$}{
            $L \leftarrow $ \KwSwap{$\alpha$, $\beta$}\;
        } 
    } 
} 
\BlankLine

$\#L$ denotes the number of different labels in $L$.\;
\KwSwap{$\alpha$, $\beta$}: swapping labels $\alpha$ and $\beta$ using QPBOI.\;
By letting $\beta = \#L+1$ the algorithm explore new number of clusters, this may affect the number of labels $\#L$.\;
\end{algorithm}

\subsection{Adaptive-label ICM}
Another discrete energy minimization method that we modified to cope with the three challenges of the CC optimization is ICM \cite{Besag1986}.
It is a point-wise greedy search algorithm.
Iteratively, each variable is assigned the label that minimizes the energy, conditioned on the current labels of all the other variables.
ICM is commonly used for MAP estimation of energies with a {\em fixed} number of labels.
Here we present an {\em adaptive-label ICM}: using the ICM conditional iterations
we adaptively determine the number of labels $k$.
Conditioned on the current labeling, we assign each point to the cluster it is most attracted to, or to a singleton cluster if it is repelled by all.

\ 

In this section we proposed a new perspective on CC optimization.
Interpreting it as MAP estimation of Potts energy allows us to propose a variety of efficient optimization methods\footnote{Matlab implementation available at: \protect\url{http://www.wisdom.weizmann.ac.il/~bagon/matlab.html}.}:\\*
\noindent\textbullet \ Swap-and-Explore (with binary step using QPBOI)\\*
\noindent\textbullet \ Expand-and-Explore (with binary step using QPBOI)\\*
\noindent\textbullet \ Adaptive-label ICM

%
Our proposed approach has the following advantages:\\*
\noindent(i)~It solves only for $n$ integer variables.
This is by far less than the number of variables required by existing methods described in Sec.~\ref{sec:existing-cc-optimization}, that requires $\sim n^2$ variables of the adjacency matrix $X=UU^T$. It makes our approach capable of dealing with large number of variables ($>100K$) and suitable for pixel-level image segmentation.\\*
\noindent(ii)~The algorithms solve directly for the cluster membership of each point, thus there is no need for rounding scheme to extract $U$ from the adjacency matrix $X$.\\*
\noindent(iii)~The number of clusters $k$ is optimally determined by the algorithm and it does not have to be externally supplied like in many other clustering/segmentation methods.

In their work Elsner and Schudy \shortcite{elsner2009} proposed a greedy algorithm to optimize the CC functional over complete graphs. Their algorithm is in fact an ICM method presented outside the proper context of CRF energy minimization, and thus does not allow to generalize the concept of discrete optimization to more recent optimization methods.

%

%
\section{Experimental Results \label{sec:synthetic}}

This section evaluates the performance of our proposed optimization algorithms using both synthetic and real data.
We compare to both existing discrete optimization algorithms that can handle multi-label non sub-modular energies (TRW-S \cite{kolmogorov2005} and BP \cite{Pearl1988}\footnote{Since these two algorithms work only with pre-defined number of clusters $k$, we over-estimate $k$ and report only the number of {\em non empty} clusters in the solution.}), and to existing state-of-the-art CC optimization method of Vitaladevuni and Basri \shortcite{Vitaladevuni2010}.
Since existing CC optimization methods do not scale beyond several hundreds of variables, extremely small matrices are used in the following experiments.
We leave it to Sec.~\ref{sec:results} to evaluate our method on large scale problems.

\subsection{Synthetic data}
\begin{figure*}
\centering
\hspace*{-1cm}
\begin{tabular}{cccc}
(a) Energy (lower=better) & (b) Recovered $k$ (GT in dashed) & (c) Purity & (d) Run time \\
\includegraphics[width=.25\linewidth]{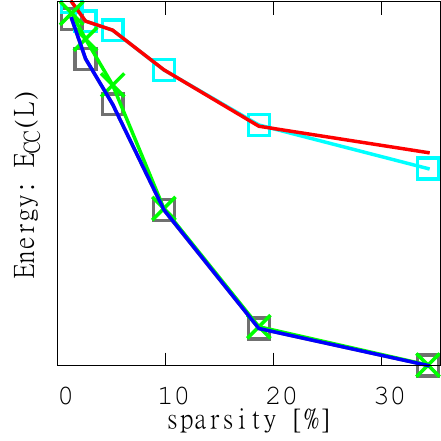}&
\includegraphics[width=.25\linewidth]{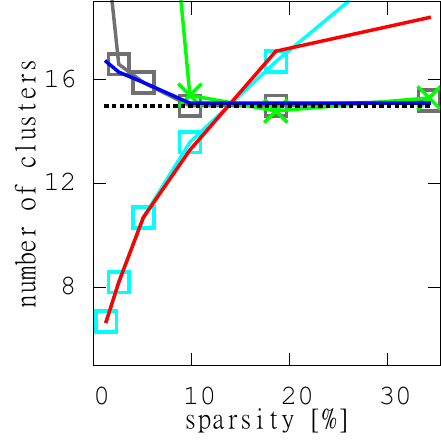}&
\includegraphics[width=.25\linewidth]{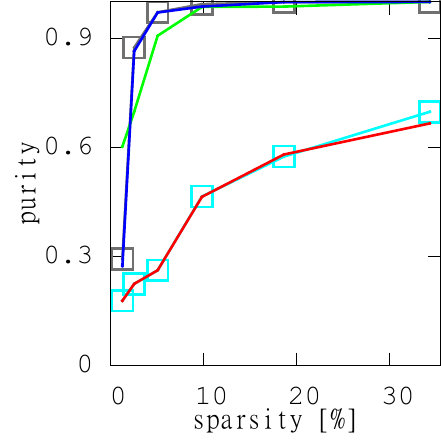}&
\includegraphics[width=.25\linewidth]{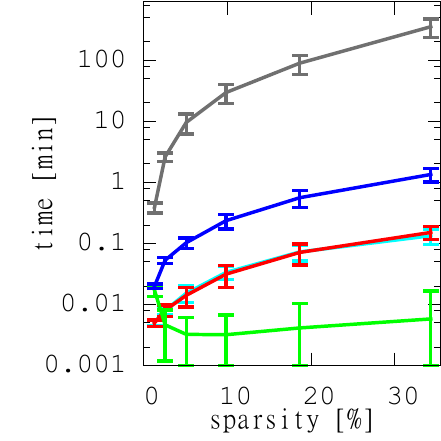}\\
\multicolumn{4}{c}{Legend:{\color{swap} Swap-and-Explore}, {\color{expand} Expand-and-Explore}, {\color{icm} ICM}, {\color{trws} TRW-S}, {\color{bp} BP}
}
\end{tabular}
\caption{{\bf Synthetic results: }{\em Graphs comparing (a)~Energy at convergence. (b)~Recovered number of clusters. (c)~Purity of resulting clusters. (d)~Run time of algorithms (log scale).
{\color{trws}TRW-S} and {\color{bp}BP} are almost indistinguishable, as are {\color{swap}Swap} and {\color{expand}Expand} in most of the plots.
\label{fig:synth-res}}}
\end{figure*}

This experiment uses synthetic affinity matrices $W$ to compare our algorithms to existing Potts optimization algorithms.
The synthetic data have 750 variables randomly assigned to 15 clusters with different sizes
(ratio between larger to smaller cluster: $\sim\times5$). For each variable we sampled roughly the same number of neighbors: of which $\sim25\%$ are from within the cluster and the rest from the other clusters.
We corrupted the clean ground-truth adjacency matrix with $20\%$ noise affecting both the sign of $W_{ij}$ and the certainty (i.e., $\left|W_{ij}\right|$). Overall the resulting percent of positive (sub-modular) connections is $\sim30\%$.

We report several measurements for these experiments: run-time, energy ($\mathcal{E}_{CC}$), purity of the resulting clusters and the recovered number of clusters $k$ for each of the algorithms as a function of the sparsity of the matrix $W$, i.e., percent of non-zero entries. Each experiment was repeated $10$ times with different randomly generated matrices.

Fig.~\ref{fig:synth-res} shows results of the synthetic experiments.
Existing multi-label approaches ({\color{trws}TRW-S} and {\color{bp}BP}) do not perform too well: higher $\mathcal{E}_{CC}$, lower purity and incorrect recovery of $k$.
This demonstrates the difficulty of the energy minimization problem that has no unary term and many non sub-modular pair-wise terms. These results are in accordance with the observations of Kolmogorov and Wainwright \shortcite{kolmogorov2005} when the energy is hard to optimize.

For our large move making algorithms, {\color{expand}Expand-and-Explore} provides marginally better clustering results than the {\color{swap}Swap-and-Explore}. However, its relatively slow running time makes it infeasible for large CC problems\footnote{This difference in run time between  Expand and Swap can be explained by looking at the number of variables involved in each of the binary steps carried out: For the expand algorithm, each binary step involves almost all the variables, while the binary swap move considers only a small subset of the variables.}.
A somewhat surprising result of these experiments shows that for matrices not too sparse (above $10\%$), {\color{icm} adaptive-label ICM} performs surprisingly well. In fact, it is significantly faster than all the other methods and manages to converge to the correct number of clusters with high purity and low energy.

From these experiments we conclude that {\color{swap}Swap-and-Explore} (Alg.~\ref{alg:ab-swap}) is a very good choice of optimization algorithm for the CC functional. However, when the affinity matrix $W$ is not too sparse, it is worth while giving our {\color{icm}adaptive-label ICM} a shot.

\subsection{Co-clustering data}

The following experiment compares our algorithms with a state-of-the-art CC optimization method, R-LP, of Vitaladevuni and Basri \shortcite{Vitaladevuni2010}.
For this comparison we use affinity matrices computed for co-segmentation.
The co-segmentation problem can be formulated as a correlation clustering problem with super pixels as the variables \cite{Glasner2011}.

We obtained 77 affinity matrices, courtesy of Glasner~\etal \shortcite{Glasner2011}, used in their experiments.
The number of variables in each matrix ranges from 87 to 788,
Their sparsity (percent of non-zero entries) ranges from $6\%$ to $50\%$,
and there are roughly the same number of positive (sub-modular) and negative (non sub-modular) entries.

Table~\ref{tab:comp-daniel} shows the ratio between our energy and the energy of R-LP method. The table also shows the percent of matrices for which our algorithms found a solution with lower energy than R-LP.
The results show the superiority of our algorithms to existing multi-label energy minimization approaches (TRW-S and BP). Furthermore, it is shown that our methods are in par with existing state-of-the-art CC optimization method on small problems.
However, unlike existing methods, our algorithms can be applied to problems {\em two orders of magnitude larger}.
Optimizing directly for $U$ not only did not compromise the performance of our method, but also allows us to handle large scale CC optimization, as demonstrated in the next section.

\begin{table}
\hspace*{-3mm}
\begin{tabular}{c||c|c|c||c|c}
&\multicolumn{3}{c||}{Ours}& & \\
& Swap & Expand & ICM & TRWS & BP  \\
\hline \hline
Energy ratio & $98.6$   & $98.4$   & $77.4$    & $83.8$   &  $83.6$  \\
 (\%)        & $\pm1.4$ & $\pm1.9$ & $\pm23.9$ & $\pm5.4$ & $\pm6.3$  \\
\hline
Strictly lower& \multirow{2}{*}{15\%} &   \multirow{2}{*}{11.7\%} & \multirow{2}{*}{0} & \multirow{2}{*}{0} & \multirow{2}{*}{0} \\
$\left(>100\%\right)$ & & & & &\\
\end{tabular}
\caption{{\bf Comparison to Glasner~\etal\ \protect\shortcite{Glasner2011}:}
{\em Ratio between our energy and of Glasner~\etal:
Since all energies are negative, higher ratio means lower energy.
Ratio higher than $100\%$ means our energy is better than Glasner~\etal.
Bottom row shows the percentage of cases where each method got strictly lower energy than Glasner~\etal.}}
\label{tab:comp-daniel}
\end{table}

%

%
\section{\label{sec:results}New Applications}

\begin{figure}
\centering
\begin{tabular}{p{.45\linewidth}p{.45\linewidth}}
\includegraphics[width=\linewidth]{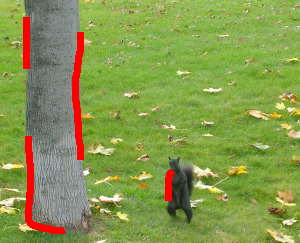}&
\includegraphics[width=\linewidth]{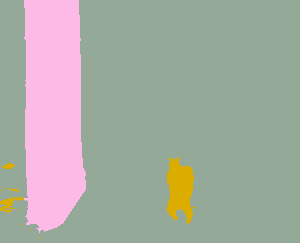}\\
\TabCenter{\linewidth}{(a) Input image and boundary scribbles (red)}&\TabCenter{\linewidth}{(b) Resulting segmentation}
\end{tabular}
\caption{\label{fig:interactive_result006}
{\bf Interactive multi-object segmentation:}
{\em (a)~The user provides only crude and partial indications to the locations of boundaries between the relevant objects in an image (red). (b)~The output of our algorithm correctly segments the image into multiple segments. Image was taken from \protect\cite{alpert2007}.}
}\vspace*{-5mm}
\end{figure}
\begin{figure*}
\centering
\includegraphics[width=\linewidth]{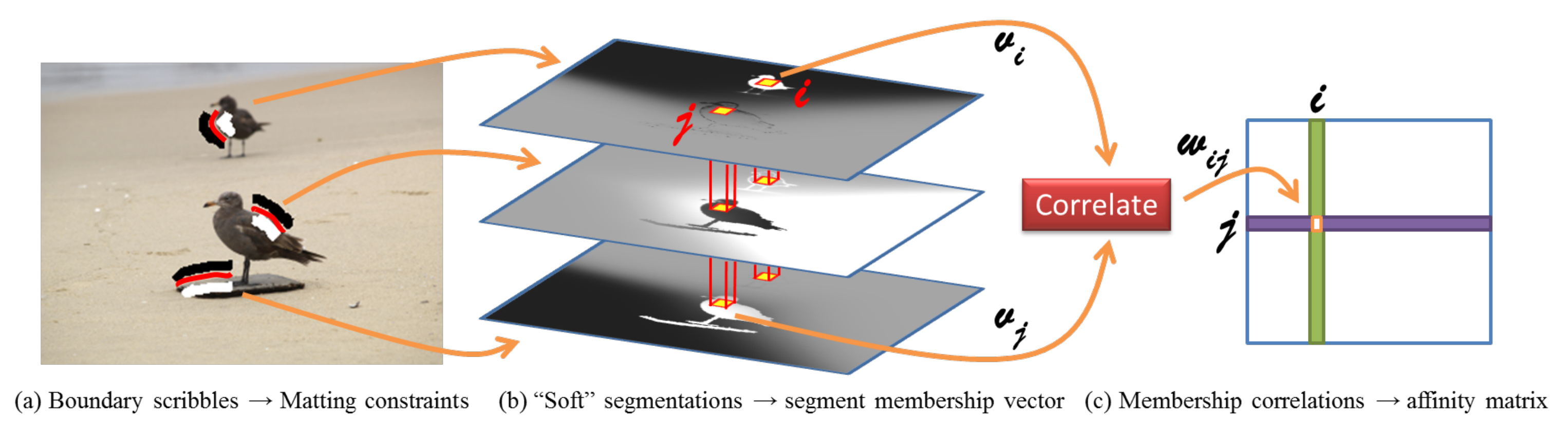}
\caption{\label{fig:neg_aff_ala_stein}
{\bf From boundary scribbles to affinity matrix:}
{\em (a)~A boundary scribble is drawn by the user (red), inducing ``figure/ground" regions on its opposite sides (black and white regions). (b)~For each scribble we use the method of Levine~\etal\ \protect\shortcite{Levin2008} to generate a soft segmentation of the image into two segments: pixel values in the soft segmentation are in the range $\left[-1,1\right]$. Pixels far away from the scribble are assigned 0 as it is uncertain to what segment they should belong to. Each pixel $i$ is described using a segmentation membership vector $v_i$ with an entry corresponding to its assignment at each soft segmentation (red columns). (c)~A non-zero entry $w_{ij}$ in the sparse affinity matrix is the {\em correlation} between normalized vectors $v_i$ and $v_j$: $w_{ij}=v_i^Tv_j/\norm{v_i}\cdot\norm{v_j}$. We also add strong repulsion across each scribble.}
}
\vspace*{-3mm}
\end{figure*}

In this section we present two new applications made possible by our large scale CC optimization.
Both these applications build upon integrates attraction and repulsion information
between large number of points, and requires the robust recovery of the underlying number of clusters $k$.


\subsection{Interactive multi-object segmentation {\color{red}(Patent Pending)}}

Our first experiment demonstrates the ability of our algorithm to handle large scale CC problem (pixel-level segmentation).

Negative affinities in image segmentation may come very naturally from boundary information: pixels on the same side of a boundary are likely to be in the same segment (attraction), while pixels on opposite sides of a boundary are likely to be in different segments (repulsion). We use this observation to design a novel approach to interactive multi-object image segmentation. Instead of using $k$ different ``strokes" for the different objects (e.g., Santner~\etal\ \shortcite{Santner2011}), the user applies a {\em single} ``brush" to indicate parts of the boundaries between the different objects. Using these {\em sparse and incomplete} boundary hints we can correctly complete the boundaries and extract the desired number of segments. Although the user does not provide at any stage the number of objects $k$, our method is able to automatically detect the number of segments using only the {\em incomplete} boundary cues.
Fig.~\ref{fig:interactive_result006} provides an example of our novel interactive multi-object segmentation approach.

\noindent{\bf Computing affinities:} Fig.~\ref{fig:neg_aff_ala_stein} illustrates how we use sporadic user-provided boundary cues to compute a {\em sparse} affinity matrix with both positive and negative entries.
Note that this is a modification of the affinity computation presented by Stein~\etal \shortcite{Stein2008}: (i)~We use the interactive boundary cues to drive the computation, rather than some boundaries computed by unsupervised technique. (ii)~We only compute a small fraction of all entries of the matrix, as opposed to the full matrix of Stein~\etal (iii)~Most importantly, we end up with both positive and negative affinities in contrast to Stein~\etal\ who use only positive affinities.

The sparse affinity matrix $W$ is very large ($\sim100k\times100k$). Existing methods for optimizing the correlation clustering functional are unable to handle this size of a matrix.
We applied our Swap-and-Explore algorithm (Alg.~\ref{alg:ab-swap}) to this problem and it provides good looking results with only several minutes of processing per image.

\begin{figure*}
\newlength{\imh}
\setlength{\imh}{1.3cm}
\centering
\hspace*{-5mm}
\begin{tabular}{c|c|c|c|c|c|c|c|c}
\includegraphics[height=\imh]{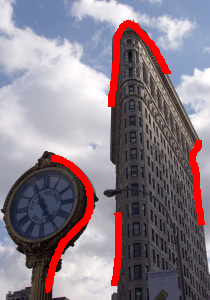}&
\includegraphics[height=\imh]{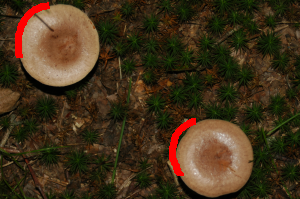}&
\includegraphics[height=\imh]{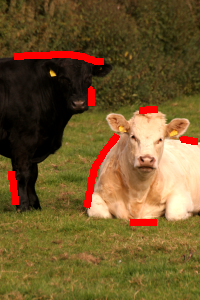}&
\includegraphics[height=\imh]{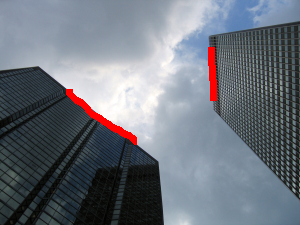}&
\includegraphics[height=\imh]{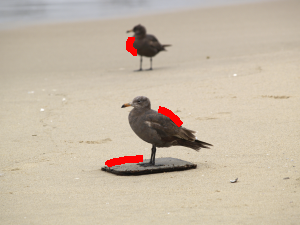}&
\includegraphics[height=\imh]{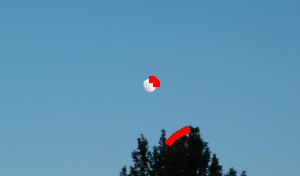}&
\includegraphics[height=\imh]{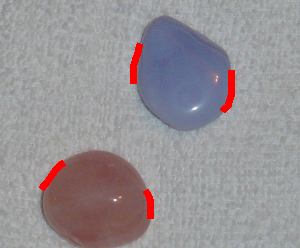}&
\includegraphics[height=\imh]{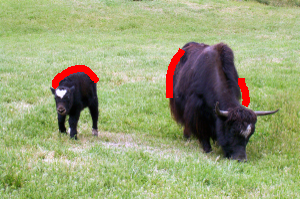}&
\includegraphics[height=\imh]{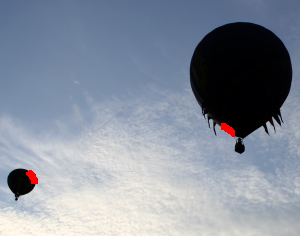}\\
\includegraphics[height=\imh]{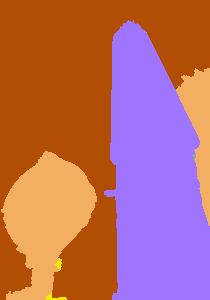}&
\includegraphics[height=\imh]{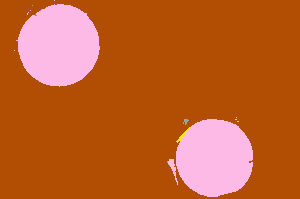}&
\includegraphics[height=\imh]{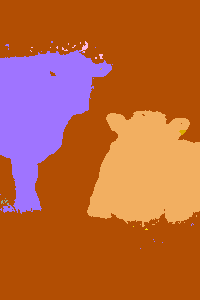}&
\includegraphics[height=\imh]{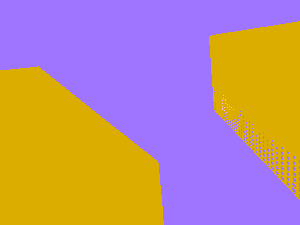}&
\includegraphics[height=\imh]{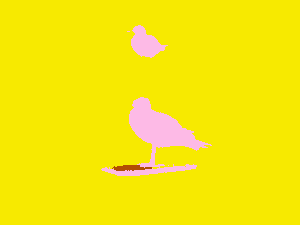}&
\includegraphics[height=\imh]{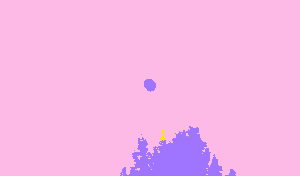}&
\includegraphics[height=\imh]{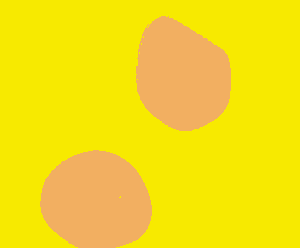}&
\includegraphics[height=\imh]{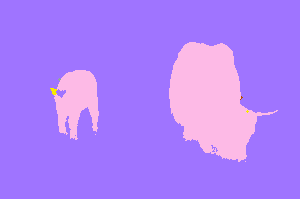}&
\includegraphics[height=\imh]{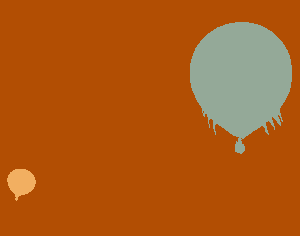}
\end{tabular}
\caption{ {\bf Interactive segmentation results. }{\em  Input image and user boundary cues (top), our result (bottom). Images were taken from \protect\cite{alpert2007}.} }
\label{fig:uimos}
\vspace*{-.3cm}
\end{figure*}

Fig.~\ref{fig:uimos} shows input images and user marked boundary cues used for computing the affinity matrix. Our results are shown at the bottom row.

The new interface allows the user to segment the image into several coherent segments without changing brushes and without explicitly enumerate the number of desired segments to the algorithm.

\subsection{Clustering and face identification}\label{sec:results-faces}
%
%

Our next experiment is to show that detecting the underlying number of clusters $k$ can be an important task on its own.
Given a collection of face images we expect the different clusters to correspond to different persons. Identifying the different people requires not only high purity of the resulting clusters but more importantly to {\em  correctly discover the appropriate number of clusters}.
This experiment is an extension of existing work on the problem of ``same/not-same" learning. Following recent metric learning approach (e.g., \cite{Guillaumin2009,Guillaumin2010}) we learn a {\em single}
classifier that assigns a probability for each pair of faces: ``how likely is this pair to be of the same person".
Then, using this classifier, we are able to determine {\em the number of persons} and cluster the faces of {\em unseen people}. That is, given a new set of face images of several {\em unseen} people, our clustering approach is able to automatically cluster and identify how many different people are in the new set of face images of {\em never seen before} people.

For this experiment we use PUT face dataset \cite{Kasinskiput2008}. The dataset consists of 9971 images of 100 people (roughly 100 images per person). Images were taken in partially controlled illumination conditions over a uniform background. The main sources of face appearance variations are changes in head pose, and facial expression.

We use the same method as Guillaumin~\etal\ \shortcite{Guillaumin2009} to describe each face. SIFT descriptors are computed at fixed points on the face at multiple scales. We use the annotations provided in the dataset to generate these keypoints.
Given a training set of labeled faces $\left\{x_i,y_i\right\}_{i=1}^N$ we use a state-of-the-art method by Guillaumin~\etal\ \shortcite{Guillaumin2010} to learn a Mahalanobis distance $L$ and threshold $b$ such that:
\[
\hspace*{-3mm}
Pr\left(y_i=y_j\vert x_i,x_j;L,b\right)=\sigma\left(b-\left(x_i-x_j\right)^TL^TL\left(x_i-x_j\right)\right)
\]
where $\sigma(z)=(1-e^{-z})^{-1}$ is the sigmoid function.

For each experiment we chose $k$ people for test (roughly $100\cdot k$ images), and used the images of the other $100-k$ people for training.
The learned distance is then used to compute $p_{ij}$, the probability that faces $i$ and $j$ belong to the same person, for all pairs of face images of the $k$ people in the test set.
The affinities are set to $W_{ij}=\log\frac{p_{ij}}{1-p_{ij}}$.
We apply our clustering algorithm to search for an optimal partition, and report the identified number of people $k^\prime$ and the purity of the resulting clusters.
We experimented with $k=15, 20, \ldots, 35$. For each $k$ we repeated the experiments for several different choices of $k$ different persons.

In these settings all our algorithms performed roughly the same in terms of recovering $k$ and the purity of the resulting clustering. However, in terms of running time adaptive-label ICM completed the task significantly faster than other methods.
We compare Swap-and-Explore to two different approaches: (i)~{\em Connected components:} Looking at the matrix of probabilities $p_{ij}$, thresholding it induces $k^\prime$ connected components. Each such component should correspond to a different person. At each experiment we tried 10 threshold values and reported the best result. (ii)~{\em Spectral gap:} Treating the probabilities matrix as a {\em positive} affinity matrix we use NCuts \cite{shi2000} to cluster the faces. For this method the number of clusters $k^\prime$ is determined according to the spectral gap: Let $\lambda_i$ be the $i^{th}$ largest eigenvalue of the normalized Laplacian matrix, the number of clusters is then $k^\prime=\arg\max_i\frac{\lambda_i}{\lambda_{i+1}}$.

\begin{figure}
\centering
\includegraphics[width=.48\linewidth]{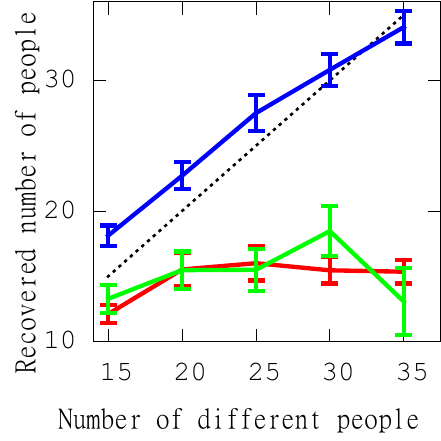}
\includegraphics[width=.48\linewidth]{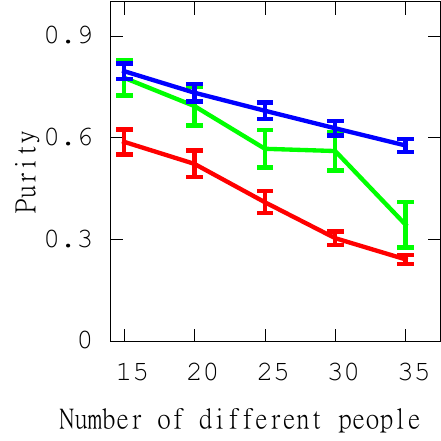}
\caption{\label{fig:PUT_faces}
{\bf Face identification:} {\em Graphs showing {\color{Blue} our result (Swap)}, {\color{Green} spectral} and {\color{Red} connected components}. Left: recovered number of people ($k^\prime$) vs. number of people in the test set. Dashed line shows the true number of people. Right: purity of resulting clusters.}}\protect\vspace*{-4mm}
\end{figure}

Fig.~\ref{fig:PUT_faces} shows cluster purity and the number of different persons $k^\prime$ identified as a function of the actual number of people $k$ for the different methods. Our method succeeds to identify roughly the correct number of people (dashed black line) for all sizes of test sets, and maintain relatively high purity values.

%

%
\section{Conclusion\label{sec:concl}}

This work provides generative probabilistic interpretation for the Correlation Clustering functional,
justifying its intrinsic ``model selection" capability.
Using a generative probabilistic formulation allows
for a better understanding of the functional,
underlying assumptions it makes, including the prior it imposes on the solution.

Apart from establishing theoretic aspects of the CC functional,
this work also suggests a new perspective on the functional, viewing it as a discrete Potts energy.
The resulting energy minimization presents three challenges: (i)~the energy is non sub-modular, (ii)~the number of clusters is not known in advance, and (iii)~there is no unary term.
We proposed new energy minimization algorithms that can successfully cope with these challenges.

Optimizing large scale CC and robustly recovering the underlying number of clusters allows us to propose new applications: interactive multi-label image segmentation and unsupervised face identification.


%

\section*{Acknowledgments}
The authors would like to thank these people for their fruitful and insightful remarks:
Ronen Basri, Michal Irani, Boaz Nadler, Shiv Vitaladevuni, Daniel Glasner, Stella Yu, Tal Hassner and Lena Gorelick.

\bibliographystyle{my_aistats}
\bibliography{negaff}

\end{document}